\documentclass[letterpaper, 10 pt, conference]{ieeeconf}  

\IEEEoverridecommandlockouts                              

\usepackage[backend=biber,
            hyperref=true,
            url=false,
            isbn=false,
            doi=false,
            backref=false,
            style=ieee,
            citestyle=numeric-comp,
            sorting=nyt,
            block=none]{biblatex}

\usepackage[hidelinks]{hyperref}
\usepackage{wrapfig}
\usepackage{times}
\usepackage{amsmath} 
\usepackage{multicol}
\usepackage{graphicx}
\usepackage{amssymb}
\usepackage{booktabs}
\usepackage{multirow}
\usepackage{subcaption}
\usepackage[export]{adjustbox}
\usepackage{wrapfig,lipsum,booktabs}
\let\labelindent\relax
\usepackage{enumitem}

\usepackage{color}
\definecolor{turquoise}{cmyk}{0.65,0,0.1,0.3}
\definecolor{purple}{rgb}{0.65,0,0.65}
\definecolor{dark_green}{rgb}{0, 0.5, 0}
\definecolor{orange}{rgb}{0.8, 0.6, 0.2}
\definecolor{red}{rgb}{0.8, 0.2, 0.2}
\definecolor{darkred}{rgb}{0.6, 0.1, 0.05}
\definecolor{blueish}{rgb}{0.0, 0.3, .6}
\definecolor{light_gray}{rgb}{0.7, 0.7, .7}
\definecolor{pink}{rgb}{1, 0, 1}
\definecolor{greyblue}{rgb}{0.25, 0.25, 1}

\definecolor{PZH_color}{RGB}{0, 102, 204}
\definecolor{TSH_color}{RGB}{218, 126, 63}

\newcommand{\expnumber}[2]{{#1}\mathrm{e}{#2}}
\newcommand{\citet}[1]{\cite{#1}}
\newcommand{\citep}[1]{\cite{#1}}

\title{
TrafficGen: Learning to Generate Diverse and Realistic Traffic Scenarios
}

\author{
Lan Feng\textsuperscript{$\mathsection$*},
Quanyi Li\textsuperscript{$\dagger$*},
Zhenghao Peng\textsuperscript{$\spadesuit$*},
Shuhan Tan\textsuperscript{$\ddagger$ },
Bolei Zhou\textsuperscript{$\spadesuit$}\\ %
\textsuperscript{$\mathsection$} ETH Zurich,
\textsuperscript{$\dagger$} The University of Edinburgh,\\
\textsuperscript{$\spadesuit$} University of California, Los Angeles,
\textsuperscript{$\ddagger$} The University of Texas at Austin
\thanks{Lan Feng, Quanyi Li and Zhenghao Peng contribute equally to this work.}
}

\begin{document}

\maketitle

\thispagestyle{empty}
\pagestyle{empty}

\begin{abstract}
Diverse and realistic traffic scenarios are crucial for evaluating the AI safety of autonomous driving systems in simulation. This work introduces a data-driven method called TrafficGen for traffic scenario generation. It learns from the fragmented human driving data collected in the real world and then generates realistic traffic scenarios. TrafficGen is an autoregressive neural generative model with an encoder-decoder architecture. In each autoregressive iteration, it first encodes the current traffic context with the attention mechanism and then decodes a vehicle's initial state followed by generating its long trajectory. We evaluate the trained model in terms of vehicle placement and trajectories, and the experimental result shows our method has substantial improvements over baselines for generating traffic scenarios. After training, TrafficGen can also augment existing traffic scenarios, by adding new vehicles and extending the fragmented trajectories. We further demonstrate that importing the generated scenarios into a simulator as an interactive training environment improves the performance and safety of a driving agent learned from reinforcement learning. Model and data are available at \url{https://metadriverse.github.io/trafficgen}.

\end{abstract}


\section{Introduction}
\label{sec:intro}

Autonomous driving (AD) is transforming our daily life with promised benefits like safe transportation and efficient mobility. One of the biggest hurdles for deploying AD in the real world is to ensure the vehicles controlled by algorithms operate safely and reliably in all kinds of traffic scenarios. Before the real-world deployment of AD, the simulation environment becomes an ideal testbed to evaluate the reliability and safety of AD systems. However, most of the existing simulators like CARLA~\cite{Dosovitskiy17} and SMARTS~\cite{zhou2020smarts} have hand-crafted traffic-generating rules and maps, while the traffic scenarios available for testing are also far from enough to emulate the complexity of the real world. As a result, it is difficult to evaluate how the AD systems make safe-critical decisions and react to other traffic participants in complex traffic scenarios. Thus, creating diverse and realistic traffic scenarios in simulation becomes crucial for thoroughly evaluating the AI safety of AD systems.

Existing methods tackle the challenge by replaying vehicle trajectories collected from the real world~\cite{li2021metadrive, kothari2021drivergym, vinitsky2022nocturne}.
Though the trajectories replayed from real data preserve the fidelity of the real world, there are two issues. First, it requires a time-consuming data collection process, particularly on a large scale. Second, the trajectories of vehicles in the commonly used driving datasets such as Waymo Open Dataset~\cite{waymo_open_dataset} and Argoverse~\cite{chang2019argoverse} are fragmented and incomplete. Most of the trajectories only span a short period of time due to the fact that they are collected by a data collection vehicle in moving, which is often occluded by other traffic participants. For instance, only 30\% of the trajectories collected in Waymo Motion Dataset~\cite{waymo_open_dataset} last more than 10 seconds, and only 12\% cover the whole scenario. Thus, the replayed traffic scenarios are incomplete and insufficient for a thorough evaluation of AD systems.

One solution for covering more traffic scenes is to design test cases manually~\cite{hauer2019did} or use heuristic methods like procedural generation (PG)~\cite{gambi2019automatically,li2021metadrive} to create a huge number of test cases. However, the generated scenarios cannot well reflect the complexity of real-world traffic and road structure. In addition, it requires a substantial amount of human effort and domain knowledge to design rules for placing vehicles, determining their initial states, and setting trigger conditions for assuring the interaction between the ego vehicle and other traffic participants~\cite{fremont2019scenic}.   

\begin{figure*}[!t]
\centering
\includegraphics[width=\linewidth]{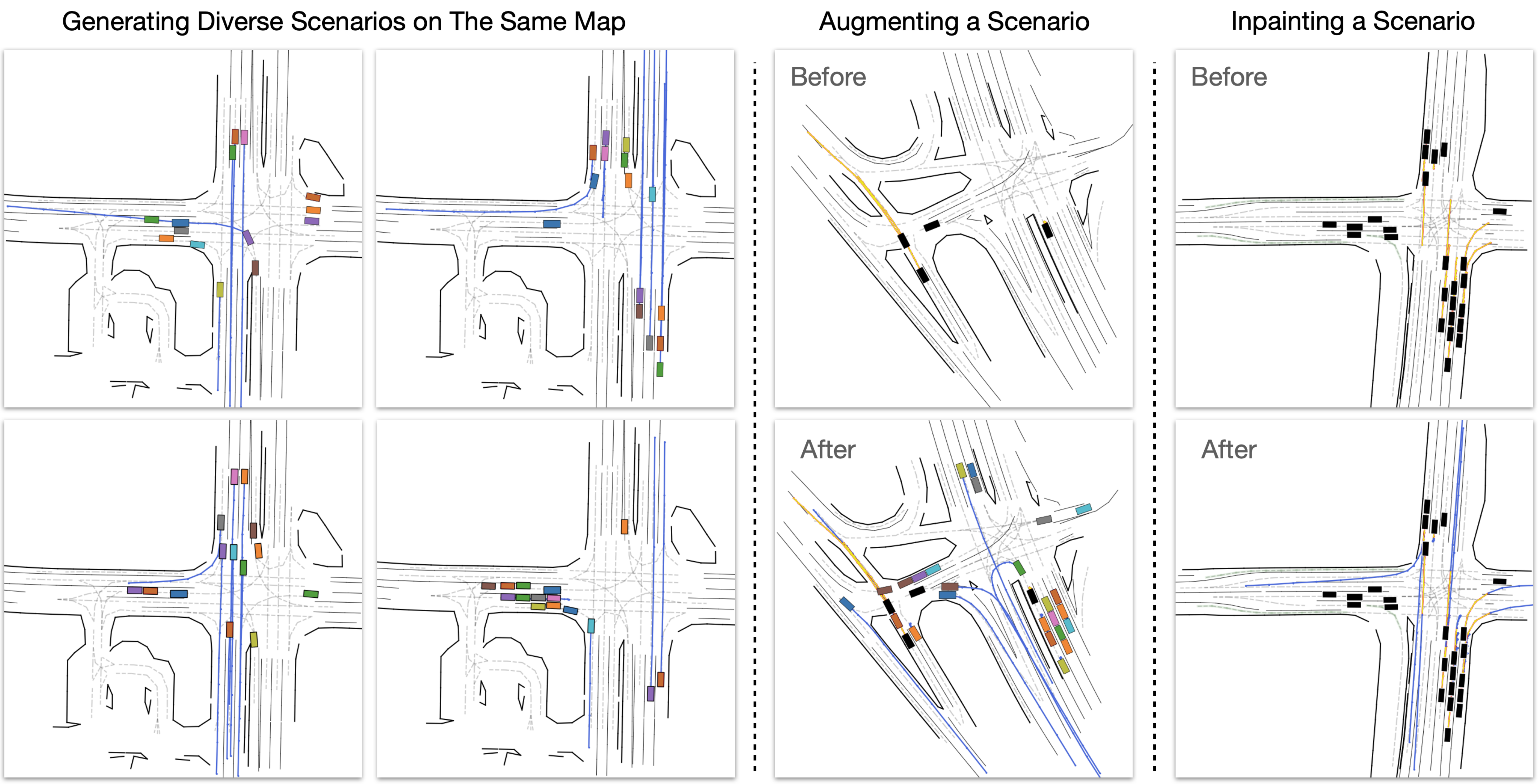}
\caption{
Different applications of TrafficGen. The colored vehicles and blue trajectories are generated by TrafficGen. The black vehicles and yellow trajectories are from the real data replay.
}
\label{fig:teaser}
\end{figure*}

Our goal is to enable the automatic generation of realistic, complete, and diverse traffic scenarios that learn from real-world data.
We develop a data-driven traffic scenario generator \textit{\textbf{TrafficGen}} that can synthesize both the initial states of traffic vehicles and their long and complete trajectories. 
TrafficGen learns from the fragmented and noisy trajectories collected in the real-world driving dataset such as Waymo Open Dataset~\cite{waymo_open_dataset}. 
TrafficGen follows an encoder-decoder neural architecture.
The encoder transforms the HD map and states of vehicles into a scenario representation.
The decoder generates initial state distributions for placing vehicles and long-term multi-modal probabilistic trajectories for realistic simulation.

TrafficGen can generate diverse and realistic traffic scenarios given HD maps after training, which greatly enlarges the set of traffic scenarios available for AD testing.
As shown in Fig.\ref{fig:teaser}, TrafficGen can also be used to edit and augment the existing scenarios, by (1) generating new traffic on the same HD map. (2) adding new vehicles and trajectories, (3) inpainting a trajectory segment into a longer one.
The generated scenarios are further imported into the simulator to improve the driving agent trained from reinforcement learning (RL). The experimental results show that the safety of driving agents can be substantially improved when being trained on the generated scenarios with higher complexity and traffic density.

\section{Related Work}
\label{sec:related}

The majority of the traffic scenarios in the existing driving simulators are either pre-recorded in the real world~\cite{kothari2021drivergym, bergamini2021simnet} or synthesized by hand-crafted rules~\cite{kar2019meta,SUMO2018,dosovitskiy2017carla, ding2021causalaf}, which lack sufficient diversity and realism.

To generate realistic driving scenarios, researchers attempt to apply data-driven methods to learn from a large-scale real-world dataset. 
Liang \textit{et al.}~\citet{liang2020learning} learn a probabilistic distribution over traffic flows from where new scenes can be sampled.
Tan \textit{et al.}~\citet{tan2021scenegen} propose a neural autoregressive model SceneGen, which inserts actors (vehicles) of various classes into the scene and synthesizes their sizes, orientations, and velocities. These methods model high-dimensional features of scenarios, improving the fidelity and scalability. However, they either require predefined heuristics and priors, \textit{e.g.}, scene grammars that encapsulate assumptions about the traffic rules~\cite{liang2020learning} or rely heavily on post-processing to guarantee the validity of the generated scene~\cite{tan2021scenegen}. 
For example, SceneGen needs to filter out improperly placed vehicles beyond road boundaries and can only generate traffic snapshots from an empty map. Compare to SceneGen, our method does not need post-processing and can augment existing traffic scenarios.
Moreover, all these methods can only generate snapshots of scenarios, which are static and can not be used for interactive simulation, while our approach allows dynamic traffic flows via trajectory generation.

Apart from generating static traffic snapshots, another line of works focuses on simulating realistic driving behaviors~\cite{peng2021learning,sun2021neuro,huang2021learning} and trajectories, such as TrafficSim~\citet{Suo2021TrafficSimLT} and SimNet~\citet{bergamini2021simnet}. TrafficSim only generates trajectories on real-world traffic scenarios. SimNet first generates traffic snapshots by generating bird-eye-view images of the map and then generates vehicle trajectories. Their method lacks the ability to edit the scenario. Differently, the proposed TrafficGen can edit the traffic scenario and generate the long trajectories of vehicles simultaneously. 

Thus compared to the previous works, our method is an integrated framework for generating both traffic snapshots and long vehicle trajectories, making it possible to produce countless new traffic scenarios as well as augment existing ones.

\begin{figure*}[!t]
\centering
\includegraphics[width=0.98\linewidth]{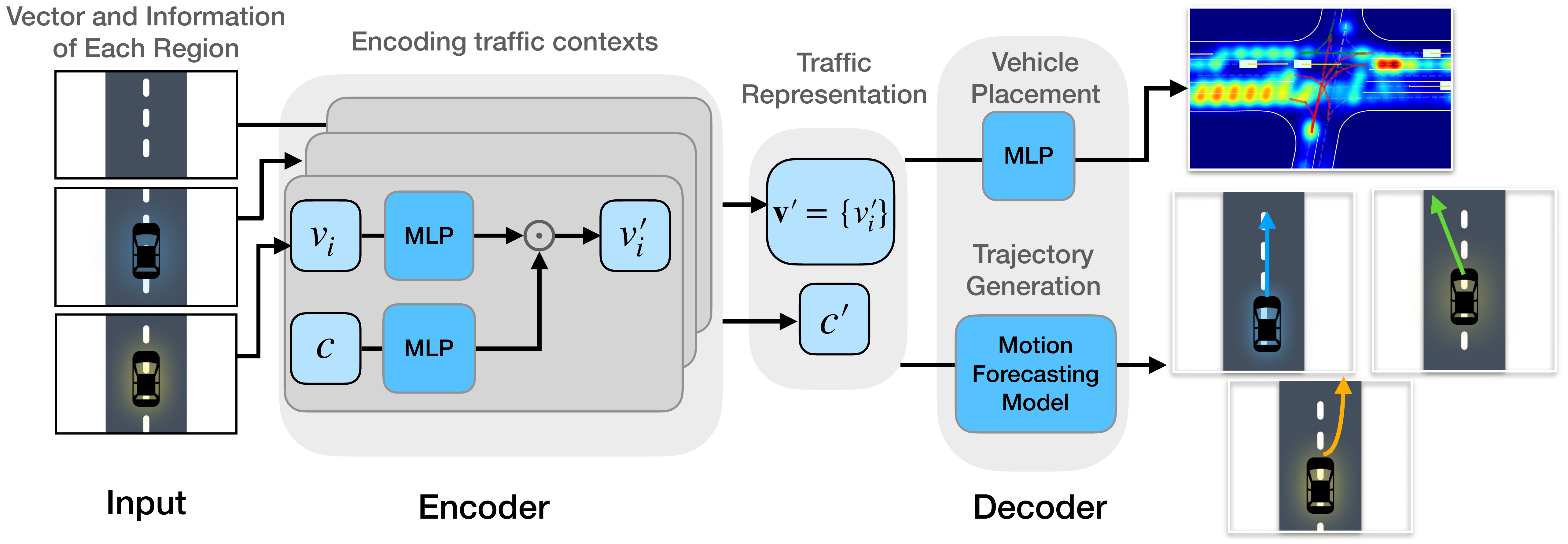}
\caption{
The neural architecture of TrafficGen. 
We first split the each HD map into $I$ regions by chunking each lane. All regions $\{v_i\}$ for each map are processed by the encoder for information fusion. The output traffic representation is then decoded to generate initial state distributions where the vehicle will be placed and the multi-modal probabilistic long-term trajectory for each vehicle.
For simplicity, we only plot one stack of encoder while we use 5 stacked blocks for encoding.
}
\label{fig:model1}
\end{figure*}
\section{Method}
\label{sec:system_design}

\noindent\textbf{Problem Formulation}.
A traffic scenario is denoted as $\tau=({\mathbf{m}, \mathbf{s}_{1:T}})$, which lasts $T$ time steps and contains the High-Definition (HD) road map $\mathbf{m}$ and the state series of traffic vehicles $\mathbf{s}_{1:T}=[s_1,...,s_T]$. Each element $s_t=\{s^1_t,...s^N_t\}$ is a set of states of $N$ traffic vehicles at time step $t$. 
Given an existing scenario $\tau=({\mathbf{m}, \mathbf{s}_{1:T}})$,
the goal of TrafficGen is to learn to generate \textbf{new} traffic scenarios $\tau{'}=({\mathbf{m}, \mathbf{s'}_{1:T'}})$ that have similar distribution with $\tau$ and different states $\mathbf{s'}$ and longer time steps $T'$. After training, TrafficGen takes $\mathbf{m}$ as input and generates $\tau'$ as a totally different scenario. Furthermore, with additional traffic fragment $[s_1,...s_t]$ as input, it can augment or inpaint the existing scenario.  
The neural architecture of TrafficGen is illustrated in Fig.~\ref{fig:model1}, where a novel encoder-decoder architecture is designed to generate traffic scenarios. We introduce each component in detail as follows.

\subsection{Encoder for Traffic Context}
\label{section:encoder-for-traffic-context}


\begin{figure}
\centering
\begin{minipage}{0.45\linewidth}
\centering
    \includegraphics[width=0.95\linewidth]{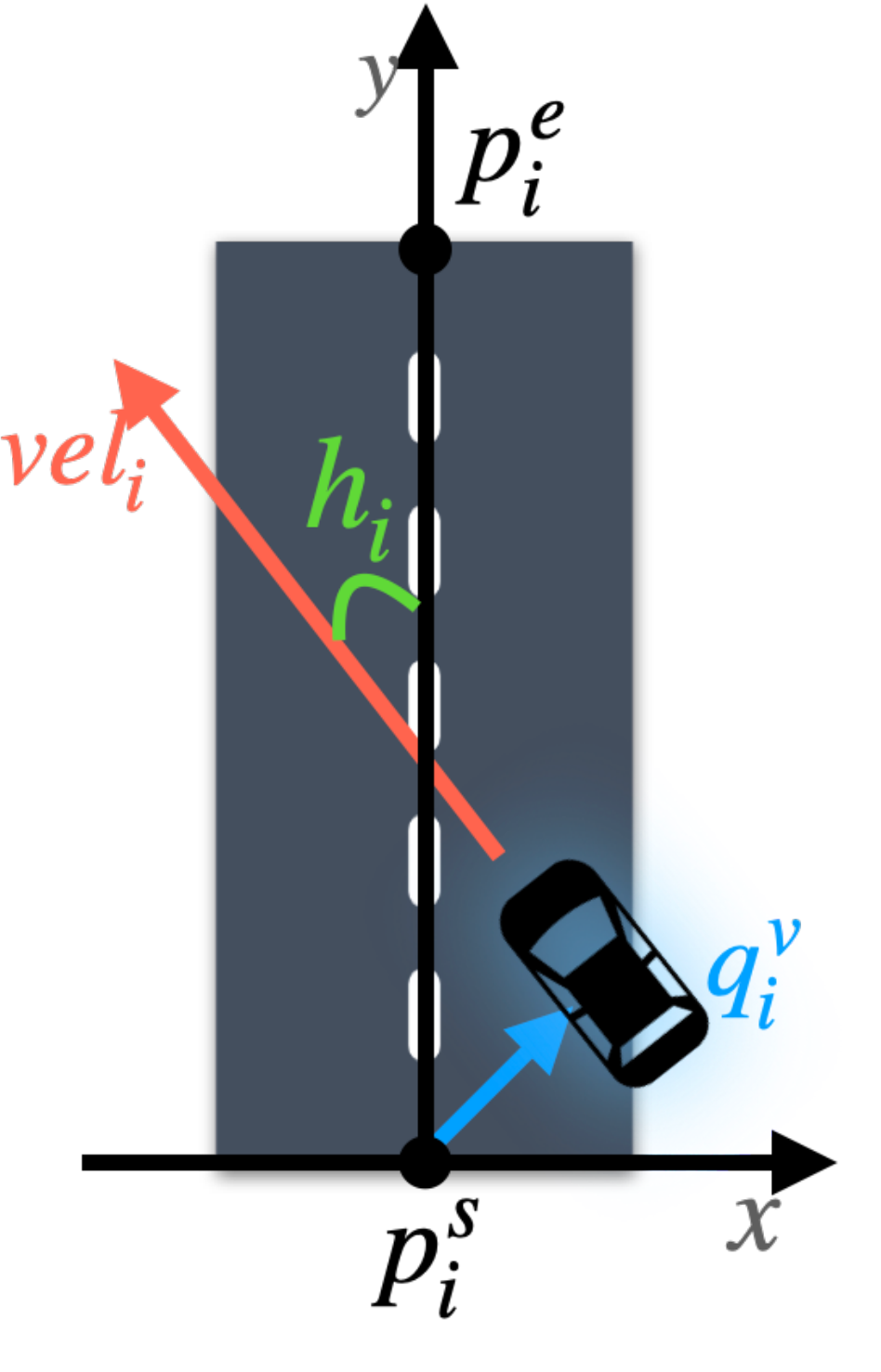}
  \caption{\centering Vector-based representation}
  \label{fig:vector_representation}
\end{minipage}
\begin{minipage}{0.4\linewidth}
\includegraphics[width=0.95\linewidth]{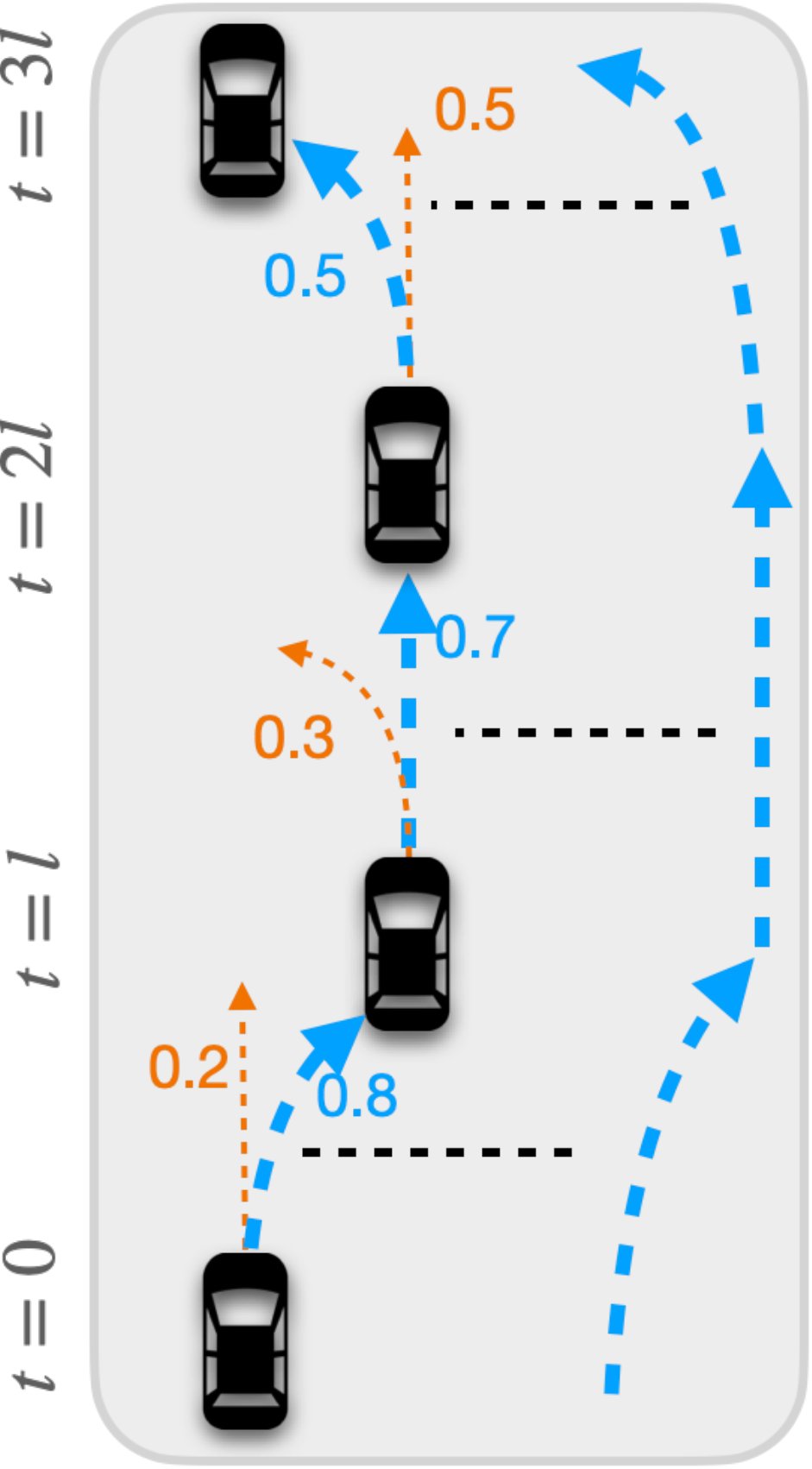}
\caption{Long trajectory generation}
\label{fig:traj_gen}
\end{minipage}
\end{figure}

\noindent\textbf{Vector-based Traffic Representation}.

We develop a new representation to encode the map and the information about all vehicles on the map.
In driving datasets like Waymo Open dataset~\cite{waymo_open_dataset} and Argoverse dataset~\cite{chang2019argoverse, wilson2021argoverse}, the map consists of a set of lanes, where each lane is described by a set of splines, consisting of an arbitrary number of 2D points connecting the center of lane lines.

In TrafficGen, we vectorize each lane into a set of vectors, each vector representing a small region in the map. Therefore the whole map can be described by a set of small regions.
For each lane, similar to the encoding in VectorNet~\cite{gao2020vectornet}, we connect two adjacent points in its center line to form a set of vectors. 
Each vector $v_i$ containing start point $p^s_{i}\in \mathbb{R}^{2}$ and end point $p^e_{i}\in \mathbb{R}^{2}$. As illustrated in Fig.~\ref{fig:vector_representation}, we establish a vector-based coordinate system for each small region whose origin is $p^s_i$ and $y$-axis directs to $p^e_i-p^s_i$. Each vector forms a rectangle region with the center as the midpoint of the vector and a side length of 5m, so that the vehicle in this region will be characterized under the local coordinates bounded by this rectangle.
Adding with the information of the vehicles in this region, the complete vector representation $v_i$ at the region bounded by $p_i^s$, $p_i^e$ at a time step is computed as: 
\begin{equation}
    v_i = (p^s_{i}, p^e_{i},t_{i},u_{i})\oplus(m_i,q^{v}_{i},h_{i},vel_{i},bbox_{i}),
\end{equation}
where $t_{i}$ represents the lane's type, and $u_{i}$ represents traffic light control state and 
$\oplus$ is concatenation. As for the state of the vehicle in this region, $m_i$ is a Boolean indicating the presence of cars in this region, 
$q^{v}_{i} \in \mathbb R^{2}$ represents the vehicle's position in the local coordinate, 
$h_{i} \in [-\frac{\pi}{2},\frac{\pi}{2}]$ is the difference between vehicle's heading and the longitudinal direction of the local coordinate, $vel_{i} \in \mathbb R$ is the vehicle's speed and $bbox_{i} \in \mathbb R^{2}$ is the vehicle's length and width. Note that if $m=0$ all the other states are also set to zero.
With the vector-based representation, we can represent a traffic snapshot $\tau$ at time step $t$ as $\tau_t=\mathbf{v}=\{v_i\}_{\ i=1}^{I}$, where $I$ is the number of vectors and $v_i$ is the information in one small region.

\noindent\textbf{Context Representation}.
Each element in $\mathbf{v}$ contains the properties of one specific region on the map. We can apply cross attention mechanism~\cite{vaswani2017attention} on the unordered set $\mathbf{v}$ to fuse the information from different regions. 
However, the size $I$ can reach up to 1000 in complex maps, which causes a huge training complexity. 
We use \textit{multi-context gating} (MCG)~\cite{varadarajan2021multipath++} to lower the training complexity because of the large number of vectors.
MCG can be viewed as an approximation to cross-attention. Instead of having each of the $n$ elements attend to all $m$ elements of the other set, MCG condenses the other set into a single context vector, denoted as $c$.

Each MCG block takes $\mathbf{v}=\{v_i\}^I$ and a context vector $c$ as input and outputs $\mathbf{v}'$
and $c'$ for the next MCG block, where $\mathbf{v}'=\{v'_i\}^I$ is the set of embeddings of each region after fusing context information and $c'$ is computed by max-pooling on $\mathbf{v}'$.
For the first MCG block, $c$ is an all-one vector.
The output features $\mathbf{v}'$, which has the same cardinality as $\mathbf{v}$, and context vector $c'$ of the last MCG block serves as the global scene representation. 


\subsection{Decoder for Traffic Scenario Generation}
\begin{figure*}[!t]
\centering
\includegraphics[width=0.95\linewidth]{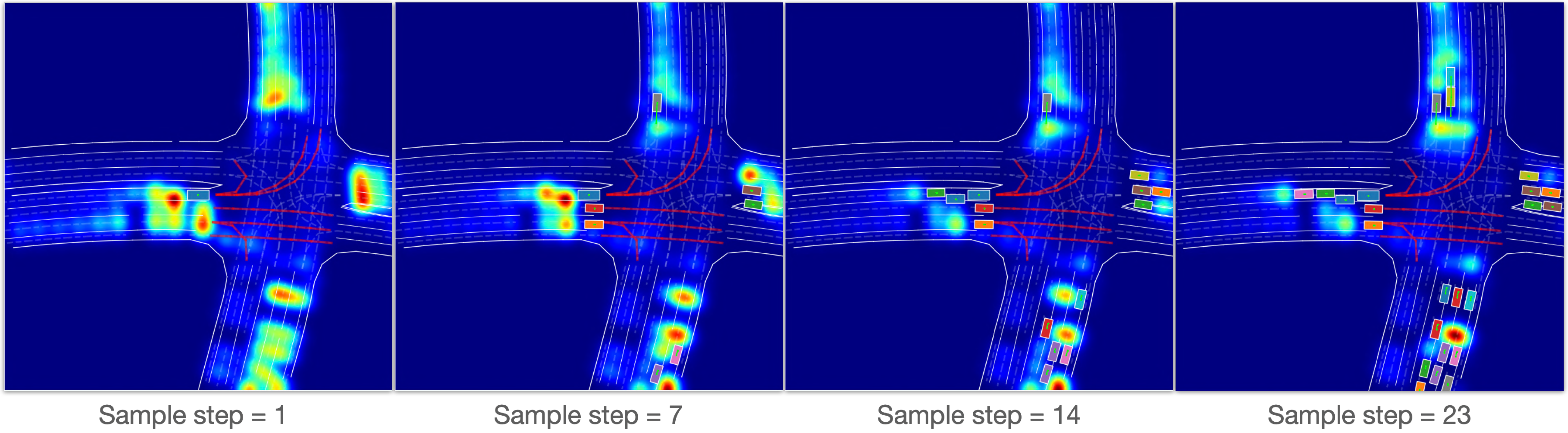}
\caption{
The autoregressive sampling process for vehicle placement. At each iteration, a vehicle is placed by sampling the spatial probability distribution over all the regions in the scene. We visualize current spatial distribution through the iteration-varying heatmap. 
At the end of sampling, a set of $N$ vehicles are placed at the map and forms a traffic snapshot.
}
\label{fig:generation}
\end{figure*}

Taking the fused feature set $\mathbf{v'}$ from the encoder and the global feature $c$ as input, the decoder of TrafficGen synthesizes a complete traffic scenario in an autoregressive way. The first step is to place vehicles on the given map and the second step is to generate their long-term trajectories.

\noindent\textbf{Vehicle Placement}.
To determine which region we should place a vehicle in, we generate a set of weights for all regions, which is later used to parameterize a categorical distribution for whether we should place a vehicle in that region: 
\begin{eqnarray}
    {w}_j = \text{MLP}_\text{place} (v'_j) \ \forall j = 1, ..., I,
    \\
    i \sim \text{Categorical}([w_1, ..., w_I]^\intercal ), \label{eq:distribution-of-presence}
\end{eqnarray}
where $i$ is the selected region. After selecting the region, we retrieve the regional feature $v'_i$ from the fused feature set $\mathbf{v'}$ and feed it to a set of networks to model the distributions of the properties of a tentative vehicle.
The distribution of its local position is modeled by a mixture of $K$ bivariate normal distributions:
\begin{eqnarray}
      [\pi_{\text{pos}}, \mu_{\text{pos}}, \Sigma_{\text{pos}}] =  \text{MLP}_\text{pos} (v'_i),
    \\
     \text{k} \sim  \text{Categorical}(\pi_{\text{pos}}),
    \\
     {q}_{i} \sim  \text{Normal}(\mu_{\text{pos, k}}, \sigma_{\text{pos, k}}).
\end{eqnarray}
Here $\mu_{\text{pos, k}}, \Sigma_{\text{pos, k}}$ represents the mean and covariant matrix for $K$ normal distributions, and $\pi_{\text{pos}}$ represents the categorical weights of the $K$ distributions. We abbreviate the resulting mixture distribution as $\text{GMM}_{\text{pos}, i}$. The log-likelihood of a local position $q_{i}$ under this distribution can be computed as $\text{GMM}_{\text{pos}, i}(q_{i})$. During inference, we can easily sample from this distribution with ${q}_{i} \sim \text{GMM}_{\text{pos}, i}$. We represent the heading, speed, and size of the vehicle in the same way. 
In the same way, we can build $K-$way mixture distributions $\text{GMM}_{\text{heading}, i}(h_i)$, $\text{GMM}_{\text{speed}, i}(v_i)$ and $\text{GMM}_{\text{size}, i}(bbox_i)$ for modeling heading, speed, and size of the car, respectively.


As shown in Fig.~\ref{fig:generation}, we use autoregressive sampling to create a traffic snapshot. We sample one vehicle with the spatial probability distribution over all the regions at a time following Eq.~\ref{eq:distribution-of-presence} and then add the sampled vehicle into that region. The attributes of the added vehicle are sampled from the attribute distributions. After that, a new traffic context is encoded and the next vehicle is generated in the next iteration. When the vehicle number reaches a predefined value, the sampling process stops.

\noindent\textbf{Training with Random Mask}. 
Inspired by seq2seq~\cite{li2018seq2seq} approaches in natural language processing, we design a scenario completion task for training the decoder. 
Similar to the training of BERT~\cite{devlin2018bert}, 
given a ground-truth traffic snapshot from the dataset, we randomly mask a number of regions containing traffic vehicles and let the model reconstruct the masked region. The decoder is trained to minimize the binary cross entropy ($BCE$) loss between the distribution in Eq.~\ref{eq:distribution-of-presence} and the ground truth Boolean indicator $m_i$. 
For the prediction of speed, heading, position, and size,
the negative log-likelihood of ground truth in the predicted mixture models is used as the loss.
Thus, the total loss can be written as:
\begin{equation}
\begin{split}
loss = & BCE({w_i},m_i)-\log\text{GMM}_{\text{pos}, i} (q_{i}) \\ & -\log\text{GMM}_{\text{heading}, i} (h_{i}) - \log\text{GMM}_{\text{speed}, i} (vel_{i})   \\ & - \log\text{GMM}_{\text{size}, i} (bbox_{i}).
\end{split}
\end{equation}

The random mask design helps the model capture traffic context information, enabling TrafficGen to edit existing traffic scenarios.





\noindent\textbf{Trajectory Generation}.
We use a motion forecasting model similar to \citet{varadarajan2021multipath++} as the multi-mode trajectory generator.
However, motion forecasting models usually take agent past trajectories as input, and cannot be directly used for long trajectory generation, since they are brittle to distributional shift.
To address these issues, we replace the past trajectory input with the global feature $c'$ and design a real-time trajectory sampling method.

As shown on the output of architecture in Fig.~\ref{fig:model1}, at each time step for each vehicle, $K$ possible $L$-step trajectories for the ego vehicle and $K$ corresponding likelihoods of them are directly generated by the decoder as
\begin{equation}
Decoder(c')= \{ pos^{t:t+L}_{k}, prob_{k} \}_{k=1}^{K}.
\end{equation}
We sample one possible future trajectory based on $K$ probabilities for each vehicle. 
To mitigate the long-term cumulative error, only the first $l$ steps of the trajectory are used, where
$l\leq L$ is a hyper-parameter that controls the update frequency. 
The long trajectories of a vehicle are generated by several rollouts of the decoder.
As an illustrative example, in Fig.~\ref{fig:traj_gen}, we show a long trajectory with $3l$ timesteps generated by rolling out the decoder for all the vehicles for $3$ times.
This decoder is trained by minimizing the MSE loss of the predicted trajectory which is the closest to the ground truth trajectory.




\section{Experiments}
\label{sect:exp}

\subsection{Experiment setting.} 

We train TrafficGen with real-world traffic scenarios imported from Waymo Open Dataset~\cite{waymo_open_dataset}. Specifically, this dataset contains about 70,000 scenarios, each with 20s trajectories. We filter out the scenarios with less than 8 agents and crop a rectangular area with a side length of 120m centered on the ego vehicle.
The rest cases are further split into a training dataset (50,000 cases) and a non-overlapping test dataset (1,000 cases).
The trained models are benchmarked on the test set to produce quantitative and qualitative results.
To train TrafficGen's vehicle placement decoder, we split each of the 20s scenarios into 10 traffic snapshots with 2s intervals and thus get $50,000 \times 10$ traffic snapshots as the training data. 
For TrafficGen's trajectory generator, the first 9s of ego trajectories from each of the 50,000 scenarios are used as training data. The feature size for the model is set to 1024. The size of the Gaussian mixture $K$ is set to 10.

\subsection{Quantitative Results of TrafficGen}
\noindent\textbf{Vehicle placement}. We conduct a quantitative experiment to show that TrafficGen does better work on vehicle placement than a competitive autoregressive method SceneGen~\cite{tan2021scenegen}. We implement
SceneGen by replacing image input with vector-based input so the inputs to both models remain the same. In this way, we can ensure the performance differences result from model design instead of the input representation.

For each scenario in the test dataset, we remove all the vehicles on the map and insert new vehicles via vehicle placement methods to compose a new snapshot to measure the difference between the vehicle attribute distributions compared to the real distribution in the dataset. 
Given two distributions $p$ and $q$, the maximum mean discrepancy 
 (MMD) measures the distribution distance between them as 
\begin{equation}
    \begin{aligned}
        \text{MMD}^2(p,q)= & \mathbb{E}_{x,x' \sim p}[k(x,x')] 
         + \mathbb{E}_{y,y'\sim q}[k(y,y')] \\
         & - 2 \mathbb{E}_{x \sim p, y \sim q}[k(x,y)]
    \end{aligned}
\label{eq:mmd}
\end{equation}
for kernel $k$. We use a Gaussian kernel with the total variation distance. 
For a pair of synthesized and original scenarios, we collect scene statistics including Pos, Heading, and Speed, and compute the MMD score between them, so the difference of attribute distributions between original and generated scenes can be measured. 
Specifically, the scene statistics are obtained by collecting attribute values from all vehicles. From these collections individual attribute values, $x$ are sampled and used in Eq.~\ref{eq:mmd}.
We then average the MMD score for each attribute across all scenarios in the test dataset to obtain the final scores presented in Table~\ref{tab:mmd}.
Compared with SceneGen, vehicle placement results generated by TrafficGen have a smaller distance to the real scenes in the dataset, indicating our method is able to generate more realistic vehicle placements.

\begin{table}[!t]
\centering
\caption{Maximum mean discrepancy (MMD) results.
}
\begin{tabular}{lcccc}\toprule
Method &Pos &Heading &Speed &Size\\\cmidrule{1-5}
SceneGen with VectorRep &0.1362 &0.1307 &0.1772 &0.1190\\
SceneGen &0.1452 &0.1387 &0.1860 &0.1286\\
\midrule
TrafficGen &\textbf{0.1192} &\textbf{0.1189} &\textbf{0.1602} &\textbf{0.0932}\\\midrule
\end{tabular}
\label{tab:mmd}
\end{table}

\noindent\textbf{Trajectory generation}.
To measure the fidelity of the trajectories generated by TrafficGen, we use the learned model to generate 9s trajectory for ego vehicle in all scenarios and use the following metrics proposed in TrafficSim~\cite{Suo2021TrafficSimLT}:
\begin{itemize}[leftmargin=*]
    \item {\textbf{Interaction Reasoning (SCR):} We use the scenario collision rate (SCR) to measure the consistency of the generated vehicle's behaviors. This is computed as the average percentage of vehicles that collides with others in each scenario. 
    Two vehicles are considered as colliding if their bounding boxes have overlapping above a predefined IOU threshold.
    }
    \item {\textbf{Scenario Reconstruction (ADE, FDE):} For each scenario sample, we use average distance error (ADE) and final distance error (FDE) at the last timestamp to measure the similarity between the generated scenario and the ground truth.}
    
\end{itemize}

We report the average metric values across all these scenarios. 

The quantitative result is in Table~\ref{tab:traj}. We test the effect of sampling intervals on model performance. When the sampling interval is 9s, the model degenerates into a traditional motion forecasting model. Our experiments show that higher sampling frequency significantly reduces the collision rate, which indicates that the real-time sampling technique can be successfully applied to a motion forecasting model to generate more interaction-aware trajectories.

\begin{table}[!t]
\centering
\caption{
Quality measurements of trajectories generated by TrafficGen.
}
\label{tab:traj}
\begin{tabular}{cccc}\toprule
\shortstack{Sampling Interval\\(s)}  & \shortstack{Mean ADE\\(m)} & \shortstack{Mean FDE\\(m)} &\shortstack{SCR\\(\%)} \\ \cmidrule{1-4}
9 &1.55 &4.62 &7.5  \\
3 &\textbf{1.54} &\textbf{4.59} &6.4  \\
1 &1.56 &4.64 &\textbf{4.9} \\\midrule
\end{tabular}
\end{table}

\subsection{Ablation study}

\begin{table}[!t]
\centering
\caption{MMD of different variants of TrafficGen (TG).}
\label{tab:ablation-study}
\begin{tabular}{lcccc}\toprule
Method &Pos &Heading &Speed &Size\\\cmidrule{1-5}
A) TG w/ vehicle masked &0.1420 &0.1375 &0.2329 &0.0987\\
B) TG w/ traffic light masked &0.1318 &0.1286 &0.1835 &0.1161\\
C) TG w/o VectorRep &0.1480 &0.1888 &0.2233 &0.0945\\
\midrule
TrafficGen &\textbf{0.1192} &\textbf{0.1189} &\textbf{0.1602} &\textbf{0.0932}\\\midrule
\end{tabular}
\end{table}

As shown in Table~\ref{tab:ablation-study}, we present the MMD results of different ablated versions of TrafficGen. 
We first mask out traffic context information to verify that TrafficGen is traffic context-aware. We mask existing vehicles (A) and traffic lights (B) separately during the sampling process. The experiment results show that both of these context information contribute to the final performance. 

We also test the proposed vector-based traffic representation in Sec.~\ref{section:encoder-for-traffic-context}. 
Aligning with Table~\ref{tab:mmd}, the vector-based traffic representation improves the performance of scenario generation methods SceneGen and TrafficGen. Thus, the vector-based traffic representation provides a good prior since it allows the decoder to produce a relative position/heading to the reference region.



\subsection{Application for Improving RL Safety}
\label{sect:generalization_exp}
We demonstrate that a safer driving agent can be obtained through reinforcement learning by training with the scenarios generated by TrafficGen than training with procedurally generated scenarios and original Waymo scenarios. 

\noindent \textbf{Datasets}.
\label{sect:dataset}
We summarize the data splits used in the experiment: 
\begin{itemize}[leftmargin=*]
\item \textit{Real Data}: the initial state and motion trajectories of traffic vehicles are collected in the real world (Waymo Open Dataset). 

\item \textit{Generated Data}: the initial state and motion trajectories of traffic vehicles are generated by TrafficGen. 

\item \textit{Augmented Data}: more vehicles are included in each scenario by setting a larger vehicle generation number $N$ in TrafficGen, leading to higher complexity and difficulty. 

\item \textit{Heuristic Data}: scenarios are generated through procedural generation (PG) and domain randomization. Traffic vehicles are spawned and destinations are assigned according to hand-crafted rules. They will be actuated by an IDM agent for speed control with a lane-changing agent for lateral control.

\item \textit{Test Set}: the cases collected in the real world while the data is not overlapped with the original Waymo training set.
\end{itemize}

Each training set contains 1,000 different traffic scenarios, while the test set contains 100 real-world scenarios where the maps and traffic data are not used in any training environments. 

\noindent \textbf{RL Environment Setting.} 
The scenarios from the Waymo Motion dataset~\cite{waymo_open_dataset} or from TrafficGen contain information on road structure and vehicle trajectories.
They are converted to interactive environments in MetaDrive simulator~\cite{li2021metadrive}. 
We train the RL controller of the ego vehicle through PPO~\cite{schulman2017proximal} on the above three training sets and evaluate them on the same test set. Experiments are conducted across 5 random seeds.
The process of importing scenarios, PPO algorithm hyperparameters, and task configuration including the observation, reward function, and termination condition follow~\cite{li2021metadrive} and are discussed in Appendix.
We evaluate each trained agent in the test set, one episode per scenario.
The ratio of the episodes where the agent reaches the destination is the \textit{success rate}. The higher the better. The \textit{safety violation} is the average episodic collision over 100 test environments. The lower the better.



\begin{table}[!t]
\centering
\caption{The test performance and the safety violation of agents trained on different datasets.
}
\label{tab:result}
\begin{tabular}{@{}ccc@{}}
\toprule
\shortstack{Training Set} & \shortstack{Success Rate $\uparrow$} & \shortstack{Safety Violation $\downarrow$}\\
\toprule
\shortstack{Real Data} &0.60 {\tiny $\pm$0.11}&3.65 {\tiny $\pm$0.71}\\\midrule
\shortstack{Heuristic Data} &0.31{\tiny $\pm$0.04} & 2.82{\tiny $\pm$0.71}\\\midrule
\shortstack{Generated Data} & \textbf{0.62} {\tiny $\pm$0.06}& 3.01 {\tiny$\pm$0.26}\\\midrule
\shortstack{Augmented Data} & 0.61 {\tiny $\pm$0.04}& \textbf{2.14} {\tiny $\pm$0.32}\\
\bottomrule
\end{tabular}%
\end{table}

\noindent \textbf{Results}.
As shown in Table~\ref{tab:result}, 
agents trained in the generated scenarios show lower safety violations and thus are safer in test environments, since these scenarios have higher difficulty and complexity than original Waymo cases.
Agents trained with generated data from TrafficGen and original Waymo dataset show better success rates than agents trained in heuristic scenarios. Therefore, our proposed scenario generation model TrafficGen shows promise in bridging the sim-to-real gap for training safe driving agents.

\section{Conclusion}
We present a data-driven traffic scenario generation method called \textit{\textbf{TrafficGen}}, which learns to synthesize traffic scenarios from the fragmented and noisy trajectories collected from the real-world driving dataset.
Based on an encoder-decoder architecture, TrafficGen first encodes the scene snapshot with its novel vector-based context representation, and then adds vehicles to the map in an autoregressive way and generates their long trajectories.
TrafficGen is flexible in terms of input. It can take empty maps or maps with existing partial traffic as input. The former allows it to produce diverse and novel traffic scenarios, while the latter allows augmenting existing ones via adding/removing vehicles and inpainting/extending fragmented trajectories. 
Experiments show that TrafficGen outperforms baselines in terms of vehicle placement and long trajectory generation.
We further demonstrate importing the synthesized scenarios into the driving simulator greatly improves the performance and the safety of the driving agent learned from reinforcement learning.

\noindent{\textbf{Acknowledgement}: The project is partially supported by Samsung Global Collaboration Award and Cisco Faculty Award.}




\printbibliography
\clearpage

\newpage
\section*{Appendix}



\section{Building Reactive Traffic Scenario}
\label{sec:turn_to_environment}
In this section, we introduce the scenario importing pipeline built upon a recent lightweight driving simulator~\cite{li2021metadrive}.
The simulator MetaDrive supports accurate physics simulation, multiple sensory inputs, and a flexible interface to customize maps and run the simulation efficiently. All of this makes MetaDrive an ideal platform for TrafficGen's scenario-importing pipeline.

\noindent \textbf{Importing Road Network}.
As discussed in 
Sec.~3
, we define a traffic scenario as a tuple of HD map and traffic flow.
The first step to instantiate a reactive driving environment is to import the road map into the simulator so that the vehicle can navigate to a specific point.

The basic road structure is represented by 3D polylines and polygons.
Since the lane centers and lane lines are represented by a list of points in Waymo Open Dataset~\cite{waymo_open_dataset}, we can approximate the lane by building a set of lane fragments and adding a rigid body and visual appearance for them. 
Therefore, each lane has its Frenet-coordinates.
The local coordinates can be used to localize vehicles, computing their distance to the left and right boundaries and determining whether a given object is on the lane.
The lane line such as yellow solid line and broken white line can also be rebuilt in the virtual world in a similar way.
Therefore contacts between vehicle and lane line can be detected to trigger some events like yielding cost or terminating the training if driving on yellow solid line.

For global routing, we build a road graph based on the connectivity information provided by Waymo dataset~\cite{waymo_open_dataset} to support route searching algorithms like BFS. With the routing and localization information, vehicles can follow the searched reference route to the destination and freely drive in the drivable region. 

\noindent \textbf{Importing Traffic Flow}.
The outputs of our model are state sequences of generated vehicles. Instead of synchronizing vehicles' positions and headings recorded in data frame-by-frame, we additionally apply a rule-based IDM policy~\cite{kesting2010enhanced} to actuate a vehicle following the generated path to its destination, enabling interactive behavior like deceleration, yielding, and emergency stop. These IDM vehicles utilize lidar to detect surrounding vehicles. If there are vehicles on the future trajectory of the target vehicle, the target vehicle will automatically determine the speed according to IDM.
Therefore, these vehicles can react to the RL agent. 

\section{Procedural Generation Baseline}
\noindent\textbf{Procedural Generation.}
MetaDrive is able to generate numerous maps through Procedural Generation (PG). Concretely, there are several basic road structures such as intersections and roundabouts, which serve as building blocks for the search-based PG algorithm. Given different random seed, the PG algorithm then randomly combine these blocks and compose different maps. Also, each building block has unique randomizable parameters such as curvature of \textit{Curve block}, so that domain randomization can be applied to generate the same road structures but in different shapes. Therefore, all generated maps will be unique, given a unique random seed.
To generate traffic flow on these procedurally generated maps, several hand-crafted rules are first executed for approximating the real-world traffic distribution and spawning traffic vehicles in the simulated world. After initialization, these vehicles are actuated by IDM policy too and will navigate to the randomly assigned destinations. 

\noindent\textbf{Experiment result.}
The agents trained in PG scenarios are benchmarked on the real-world test set. The awful test success rate implies that in spite of meticulous hand-crafted traffic scenarios, the driving skills learned in these synthetic scenes can not be generalized to real-world scenarios.
This, in turn, points out that methods building traffic scenarios from real-world data, such as \textit{TrafficGen}, is promising for bridging the sim-to-real gap.

\section{Environment Details}
\label{appendix:env_details}
In the driving task, the objective of RL agents is to steer the target vehicles with low-level continuous control actions, namely acceleration, brake, and steering.


\noindent\textbf{Observation.}
The observation of RL agents is as follows:
\begin{itemize}[noitemsep, leftmargin=2em]
  \item A 240-dimensional vector denoting the Lidar-like point clouds with $50 m$ maximum detecting distance centering at the target vehicle. 
  Each entry is in $[0, 1]$ with Gaussian noise and represents the relative distance of the nearest obstacle in the specified direction.
  \item A vector containing the data that summarizes the target vehicle's state such as the steering, heading, velocity, and relative distance to the left and right boundaries.
  \item The navigation information that guides the target vehicle toward the destination. 
  We sparsely spread a set of checkpoints, 50m apart on average, in the route and use the relative positions toward future checkpoints as additional observation of the target vehicle. 
\end{itemize}

\noindent\textbf{Reward and Cost Scheme.}
The reward function is composed of four parts as follows:
\begin{equation}
\label{eq:reward-functgion}
  R = c_{1}R_{disp} + c_{2}R_{speed} + R_{term}.
\end{equation}
The \textit{displacement reward} $R_{disp} = d_t - d_{t-1}$, wherein the $d_t$ and $d_{t-1}$ denotes the longitudinal movement of the target vehicle in Frenet coordinates of the current lane between two consecutive time steps, provides a dense reward to encourage the agent to move forward. 
The \textit{speed reward} $R_{speed} = v_t/v_{max}$ incentives agent to drive fast. $v_{t}$ and $v_{max}$ denote the current velocity and the maximum velocity ($80 \ km/h$), respectively.
We also define a sparse \textit{terminal reward} $R_{term}$, which is non-zero only at the last time step. At that step, we set $R_{disp} = R_{speed} = 0$ and assign $R_{term}$ according to the terminal state.
$R_{term}$ is set to $+10$ if the vehicle reaches the destination, $-5$ for crashing others or violating the traffic rule.
We set $c_1 = 1$ and $c_2 = 0.1$.
For measuring safety, collision with vehicles, obstacles, and sidewalk raises a cost of $+1$ at each time step. The sum of cost generated in one episode is episode cost, a metric like an episode reward, but reflecting safety instead.

\noindent\textbf{Termination Conditions and Evaluation Metrics.} 
Since we attempt to benchmark the safety of trained agents, collisions with vehicles and the sidewalk will not terminate the episode. The episode will be terminated only when: 1) the agent drives out of the drivable area, such as driving outside of the yellow solid line, 2) the agent arrives at the destination and 3) the episode length exceeds the pre-defined horizon (1000 steps). 
For each trained agent, we evaluate it in 100 held-out test environments and define the ratio of episodes where the agent arrives at the destination as the \textit{success rate}. The \textit{Episodic Cost}, also referred to as \textit{Safety Violation} is the average episode cost on 100 test environments.
Since each agent are trained across 5 random seeds, this evaluation process will be executed for 5 agent which has the same training setting but different random seeds. We report the average and std on the 2 metrics mentioned above for measuring the performance and the safety of trained agents.

\noindent\textbf{Importing Scenarios.} 
Unlike the RL agent that can select new routes in the drivable area to the destination, 
Traffic vehicles follow the logged trajectories, while IDM is used~\cite{kesting2010enhanced, kesting2007general} for control speed so that they can react to RL agent which may behave differently from the logged ego car, performing behaviors like yielding and emergency stop.

\noindent\textbf{Statistical Results Processing.}
Since each agent is trained with 5 random seeds, the evaluation process will be executed for 5 agents which have the same training setting but different random seeds. We report the mean and std of the 2 aforementioned metrics to measure the performance and the safety of trained agents.

\section{Implementation Details}

\begin{table}[ht]
\centering
\centering
\label{hyper:tt}
\caption{TrafficGen}
\begin{tabular}{@{}ll@{}}
\toprule
Hyper-parameter             & Value  \\ \midrule
Feature Size              & 1024    \\
Training epochs   & 30     \\
Data Usage for Initializer & 50,000 \\
Data Usage for Actuator & 150,000 \\
Learning Rate   & $\expnumber{3}{-4}$ \\ 
Activation Function & ``relu'' \\
MLP Layers & 3\\
\bottomrule
\end{tabular}
\end{table}

\begin{table}[ht]
\centering
\centering
\caption{PPO}
\label{hyper:ppo}
\begin{tabular}{@{}ll@{}}
\toprule
Hyper-parameter             & Value  \\ \midrule
KL Coefficient              & 0.2    \\
$\lambda$ for GAE~\citep{schulman2018highdimensional} & 0.95 \\
Discounted Factor $\gamma$   & 0.99  \\
Number of SGD epochs   & 20     \\
Train Batch Size & 30,000 \\
SGD mini-batch size & 256 \\
Learning Rate   & $\expnumber{3}{-4}$ \\ 
Clip Parameter $\epsilon$ & 0.2 \\
Activation Function & ``tanh'' \\
MLP Hidden Units & 256 \\
MLP Layers & 2\\
\bottomrule
\end{tabular}
\end{table}

\label{appendix:resource_usage}
The training is executed on servers with 8 x Nivdia 1080ti and 256 G memory.
When training PPO agents, we use RLLib~\cite{liang2018rllib} and host 16 concurrent trials on 8 x Nvidia A100 GPU. Each trial consumes 6 CPUs with 10 parallel rollout workers. 
The total memory consumption for each trial is approximately 100 G. 

For the TrafficGen model, we set the feature dimension of $\mathbf{v'}$ to be 1024, and use 3-layer MLPs with hidden dimensions of [2048, 1024, 256] for attribute modeling. During training, we train both networks with a learning rate of $3e-4$ for 30 epochs.

Note that we make two assumptions about vehicles that are driving on the road. First, the distance between a vehicle and its nearest road center lane should be limited. We set the range limit to 5 meters in experiments. Second, the angle between a vehicle's heading and the road center lane's direction should be limited. We set the angle limit to $\pm 90^{\circ}$ in experiments. These assumptions help to filter out invalid vehicles in the dataset. Besides, they give good prior information for the model, ensuring that invalid vehicles such as the out-of-road and retrograde ones will not be generated.


\end{document}